\documentclass[journal]{IEEEtai}

\usepackage[colorlinks,urlcolor=blue,linkcolor=blue,citecolor=blue]{hyperref}

\usepackage{color,array}

\usepackage{graphicx}
\usepackage{graphicx}
\usepackage{subfigure}
\usepackage{array}
\usepackage{cite}
\usepackage{amsmath,amssymb,amsfonts}
\usepackage{mathrsfs}
\usepackage{bm}
\usepackage{multirow}
\usepackage{booktabs}
\usepackage{color}
\usepackage{enumerate}
\usepackage{bbm}
\usepackage{listings}
\usepackage{algorithm}

\begin{document}

\title{ASD: Towards Attribute Spatial Decomposition for Prior-Free Facial Attribute Recognition}

\author{Chuanfei Hu, \IEEEmembership{Student Member, IEEE}, Hang Shao, Bo Dong, Zhe Wang, and Yongxiong Wang, \IEEEmembership{Member, IEEE}
\thanks{Chuanfei Hu, Zhe Wang and Yongxiong Wang are with School of Optical-Electrical and Computer Engineering, University of Shanghai for Science and Technology, Shanghai, China (e-mail: chuanfei\_hu@ieee.org, 201440049@st.usst.edu.cn, wyxiong@usst.edu.cn).}
\thanks{Hang Shao is with School of Computer Science and Engineering, Nanjing University of Science and Technology, Nanjing, China (e-mail: shaohang@njust.edu.cn).}
\thanks{Bo Dong is with Center for Brain Imaging Science and Technology, Zhejiang University, Zhejiang, China (e-mail: bodong.cv@gmail.com).}
\thanks{This paragraph will include the Associate Editor who handled your paper.}}

\markboth{Pre-print for submission}
{Chuanfei Hu \MakeLowercase{\textit{et al.}}: ASD: Towards Attribute Spatial Decomposition for Prior-Free Facial Attribute Recognition}

\maketitle

\begin{abstract}
Representing the spatial properties of facial attributes is a vital challenge for facial attribute recognition (FAR).
Recent advances have achieved the reliable performances for FAR, benefiting from the description of spatial properties via extra prior information.
However, the extra prior information might not be always available, resulting in the restricted application scenario of the prior-based methods.
Meanwhile, the spatial ambiguity of facial attributes caused by inherent spatial diversities of facial parts is ignored.
To address these issues, we propose a prior-free method for attribute spatial decomposition (ASD), 
mitigating the spatial ambiguity of facial attributes without any extra prior information. 
Specifically, assignment-embedding module (AEM) is proposed to enable the procedure of ASD, 
which consists of two operations: attribute-to-location assignment and location-to-attribute embedding.
The attribute-to-location assignment first decomposes the feature map based on latent factors, 
assigning the magnitude of attribute components on each spatial location. 
Then, the assigned attribute components from all locations to represent the global-level attribute embeddings.
Furthermore, correlation matrix minimization (CMM) is introduced to enlarge the discriminability of attribute embeddings.
Experimental results demonstrate the superiority of ASD compared with state-of-the-art prior-based methods,
while the reliable performance of ASD for the case of limited training data is further validated.
\end{abstract}

\begin{IEEEImpStatement}
Facial attribute recognition (FAR) can label the facial attributes of a person from a facial image, 
which has been widely applied to numerous real-world applications, such as facial verification and identification, face generation, and face retrieval. 
The existing FAR methods introduce some prior information to achieve the satisfactory performances.
The prior information might not be always available, resulting in the restricted application scenario of these methods.
The proposed attribute spatial decomposition (ASD) can construct a prior-free FAR method, 
which achieves superior performance without any prior information.
The experimental results verify the effect of ASD, which could fill the blank of prior-free method in FAR task.
\end{IEEEImpStatement}


\section{Introduction}

\IEEEPARstart{F}{acial} attribute recognition (FAR) aims to predict multiple biometrics from a given facial image, such as gender, mustache, lip thickness, and hair color, which has be widely contributed to numerous real-world applications, e.g., facial verification and identification \cite{Kumar2011Describable, Di2021Multi, Li2021Enhanced, Wadhawan2022Landmark}, face generation \cite{yan2016attribute2image, Liu2021Generating}, and face retrieval \cite{Chen2013Scalable, zaeemzadeh2021face}. 
However, FAR is still an open issue in the wild, since facial appearance is variable significantly caused by complicated factors of capturing, such as illumination, pose, and occlusion.

\begin{figure}[!t]
	\centering
	\includegraphics[width=0.9\linewidth]{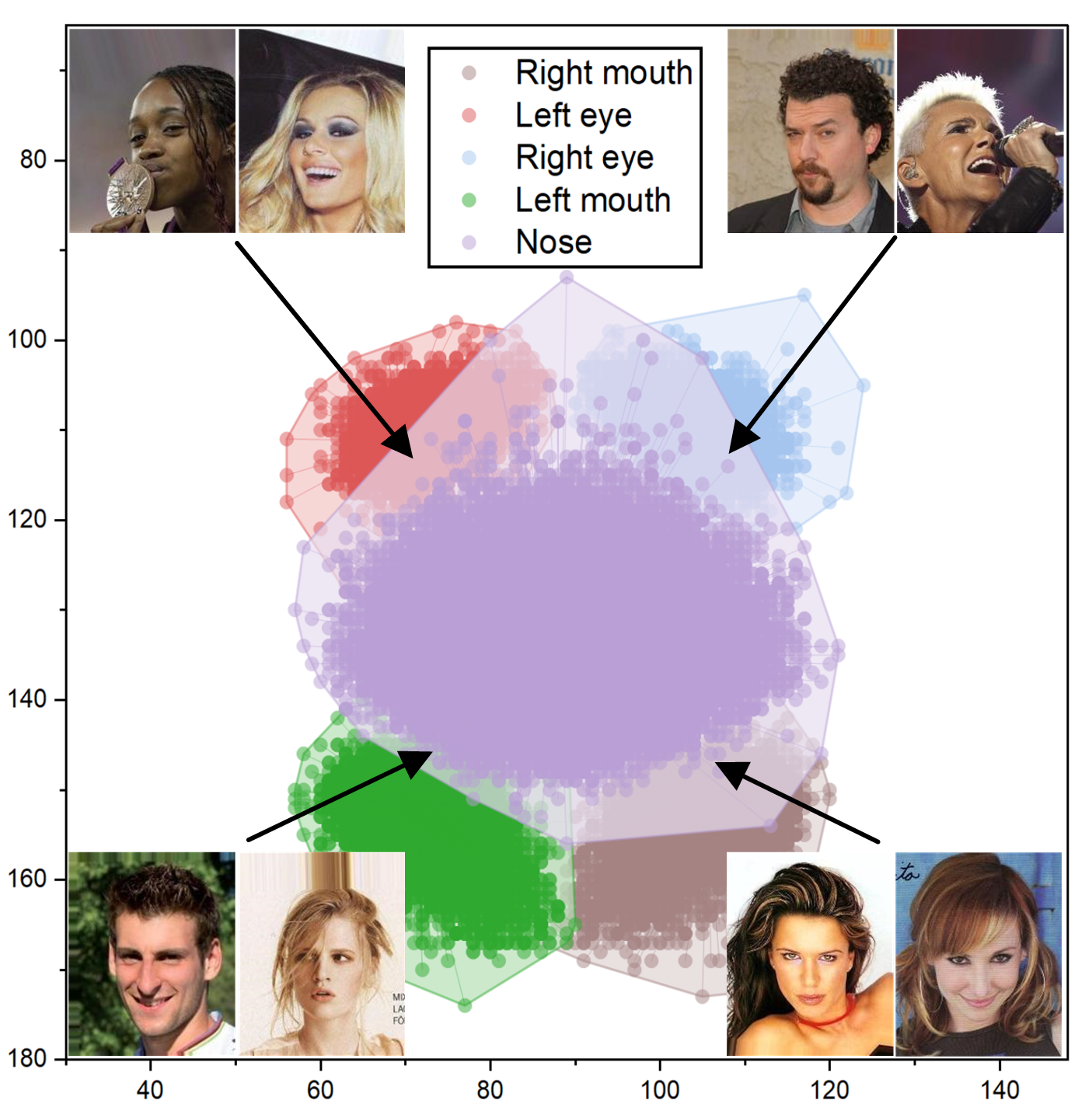}
	\caption{The spatial ambiguity of facial attributes from the statistical observation. We first plot the coordinates of the facial landmarks, including {\itshape nose}, {\itshape left eye}, {\itshape right eye}, {\itshape left mouth}, and {\itshape right mouth} for all individuals in CelebA \cite{Liu2015Deep}. Meanwhile, we further construct the convex hulls of the corresponding coordinate sets. The obvious overlapping regions among the convex hulls illustrate that a spatial location might contain the various facial attributes for different individuals, revealing the spatial ambiguity of facial attributes. }
	\label{fig:iss:a}
\end{figure}

\begin{figure}[!t]
	\centering
	\includegraphics[width=\linewidth]{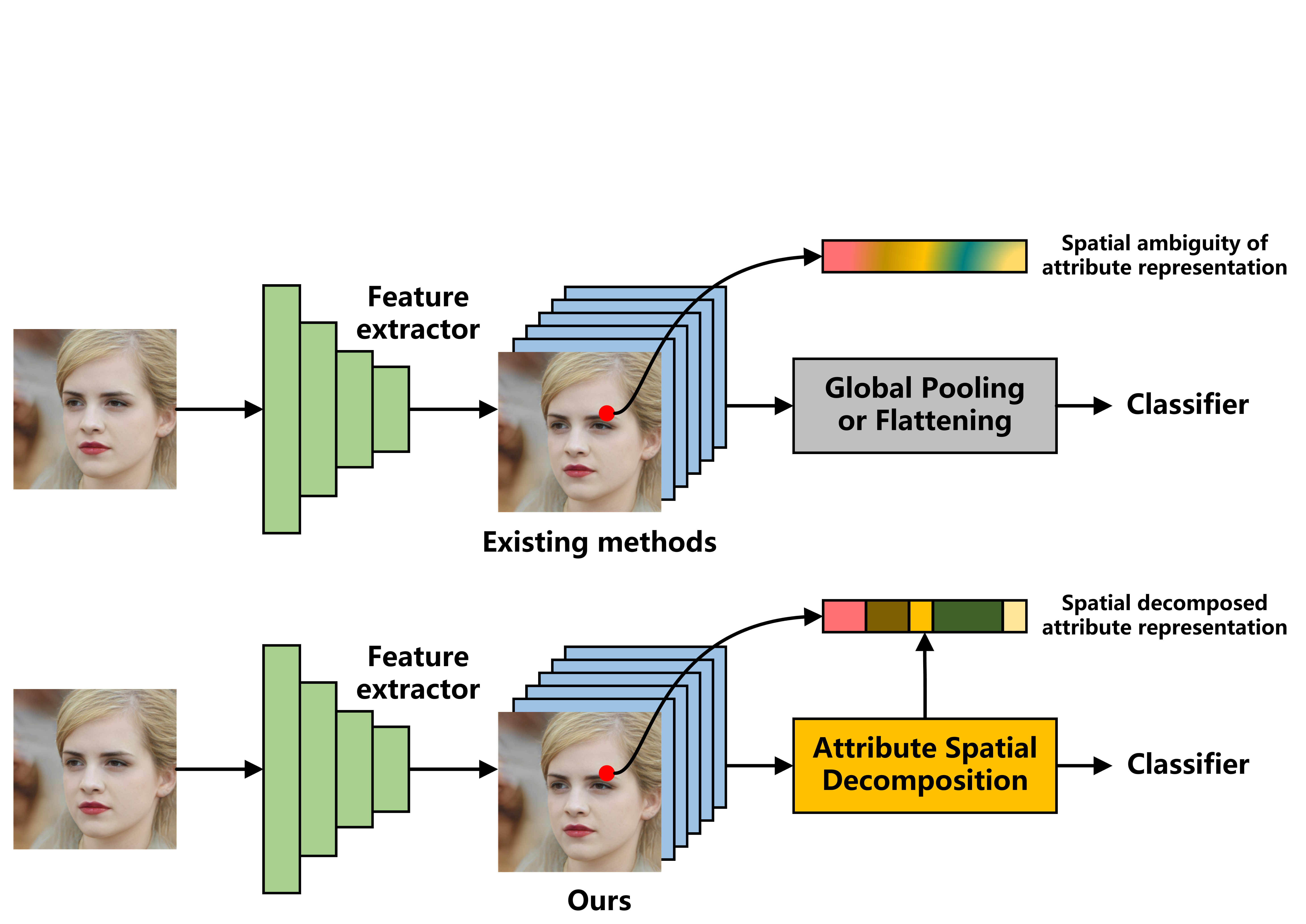}
	\caption{The insight of the proposed attribute spatial decomposition. Compared with the existing methods that obscure the spatial properties of facial attributes,
		the proposed attribute spatial decomposition (ASD) can alleviate the spatial ambiguity without any extra prior information.
	}
	\label{fig:iss:b}
\end{figure}

Recent efforts have been made towards improving deep learning-based FAR methods \cite{li2018landmark, Cao2018Partially, He2018Harnessing, mao2020deep, sharma2020slim, Shu2021Learning, Chen2021Improving} via additional annotations,
which can be seen as an insight of modeling {\itshape the spatial properties of facial attributes via extra prior information}.
These prior-based methods can be categorized as the explicit and implicit prior information-based.
The explicit prior information-based methods introduce additional annotations, such as facial landmarks \cite{mao2020deep}, identifications \cite{Cao2018Partially, Shu2021Learning}, and face parsing masks \cite{He2018Harnessing, Shu2021Learning}, to construct a multi-task learning or self-supervised learning, in which the additional annotations can be regarded as the prior information to enhance the discriminability between the features of facial attributes. On the other hand, the implicit prior information-based methods design a multi-task learning network via attribute groups \cite{sharma2020slim, li2018landmark, Cao2018Partially, Chen2021Improving}, 
which groups the attributes manually according to locations or semantics. 
Although these methods achieve the promising results in FAR, there are three major drawbacks. 
{\itshape First}, the explicit prior information might not be always available, resulting in the restricted application scenario of these methods.
{\itshape Meanwhile}, the manual grouping of facial attributes is not suitable and optimal, since different individuals might give different partitions according to locations or semantics. 
It means that the relationships among facial attributes could not be defined sufficiently via prior experiences. 
{\itshape Furthermore}, the inherent spatial diversities of facial parts, caused by different individuals and various poses, are considered insufficiently. 
For instance, we illustrate the distributions of facial landmarks, including nose, eyes, and mouth, from all individuals in CelebA \cite{Liu2015Deep}, as shown in Figure~\ref{fig:iss:a}.
The conspicuous overlapping regions of the convex hulls reveal that the spatial ambiguity of facial attributes is existed inherently. 
In the above mentioned FAR methods, flattening and global pooling are utilized to vectorize the convolutional features, in which the spatial properties are diluted, resulting in the spatial ambiguity of facial attributes, as shown in Figure~\ref{fig:iss:b}.
Therefore, a challenging issue remains: 

``{\itshape Can we design a method to alleviate the spatial ambiguity of facial attributes without any extra prior information}'' ? 

Specifically, the spatial ambiguity of facial attributes means that multiple attribute features might be represented on the same spatial location of feature map.
Intuitively, to alleviate the spatial ambiguity, conducting a classifier to recognize the attributes from each spatial location of feature map might be an available approach. 
However, the traverse of all positions would produce redundant predictions.
Since the spatial-wise labels of attributes are unavailable, we have to build a full label-space with all attributes for each spatial location, resulting in many trivial prediction results.
Furthermore, a strategy of integrating all predictions would raise the issue of increase in complexity.

In this paper, we focus on FAR method without any extra prior information, denoted as prior-free FAR method. A novel module whose capability of attribute spatial decomposition (ASD) is proposed, termed as assignment-embedding module (AEM). 
Motivated by hidden factor analysis (HFA) \cite{Gong2013Hidden}, 
a facial image can be seen as a combination of facial attribute components.
Therefore, ASD can be intuitively cast as decomposing the attribute components in the spatial dimension via latent factors, incorporated into a paradigm of supervised learning-based FAR.
This is achieved via two operations of attribute-to-location assignment and location-to-attribute embedding.
The attribute-to-location assignment aims to decompose the feature map based on latent factors, describing the attribute components on each spatial location.
Then, the location-to-attribute embedding aggregates the assigned attribute components from all locations to represent the global-level attribute embeddings.
To further enlarge the discriminability of attribute embeddings, correlation matrix minimization (CMM) is introduced to constrain the correlations among latent factors, 
muting the impact caused by relationships among latent factors on the procedure of assignment.
The advantage of ASD is to describe the attribute components formally without any extra prior information.
To summarize, the main contributions are as follows:
\begin{itemize}
	\item We propose a prior-free FAR method which can decompose the attribute components in the spatial dimension, thereby mitigating the spatial ambiguity of facial attributes.
	To the best of our knowledge, we are among the first to introduce the insight of attribute spatial decomposition (ASD) for improving the performance of FAR.
	\item A novel module, termed as assignment-embedding module (AEM), is proposed to enable the procedure of ASD via attribute-to-location assignment and location-to-attribute embedding.
	\item Correlation matrix minimization (CMM) is introduced to decorrelate the latent factors, strengthening the discriminability of the decomposed attribute embeddings.
	\item The effectiveness of ASD is demonstrated via ablation studies, 
	and the superior performances are achieved on CelebA \cite{Liu2015Deep} and LFWA \cite{Liu2015Deep} without any extra prior information.
	Moreover, the superiority of ASD adopted to the case of limited training data is validated experimentally. 
\end{itemize}

The remainder of the paper is organized as follows. 
In Section~\ref{sec:related}, we briefly describe the related works on deep learning-based FAR methods and the decomposition of facial components. 
Section~\ref{sec:method} presents the details of ASD. 
The experimental results are illustrated in Section~\ref{sec:experiment}. 
The conclusion is finally drawn in Section~\ref{sec:conclusion}.

\begin{figure*}[!ht]
	\centering
	\includegraphics[width=\linewidth]{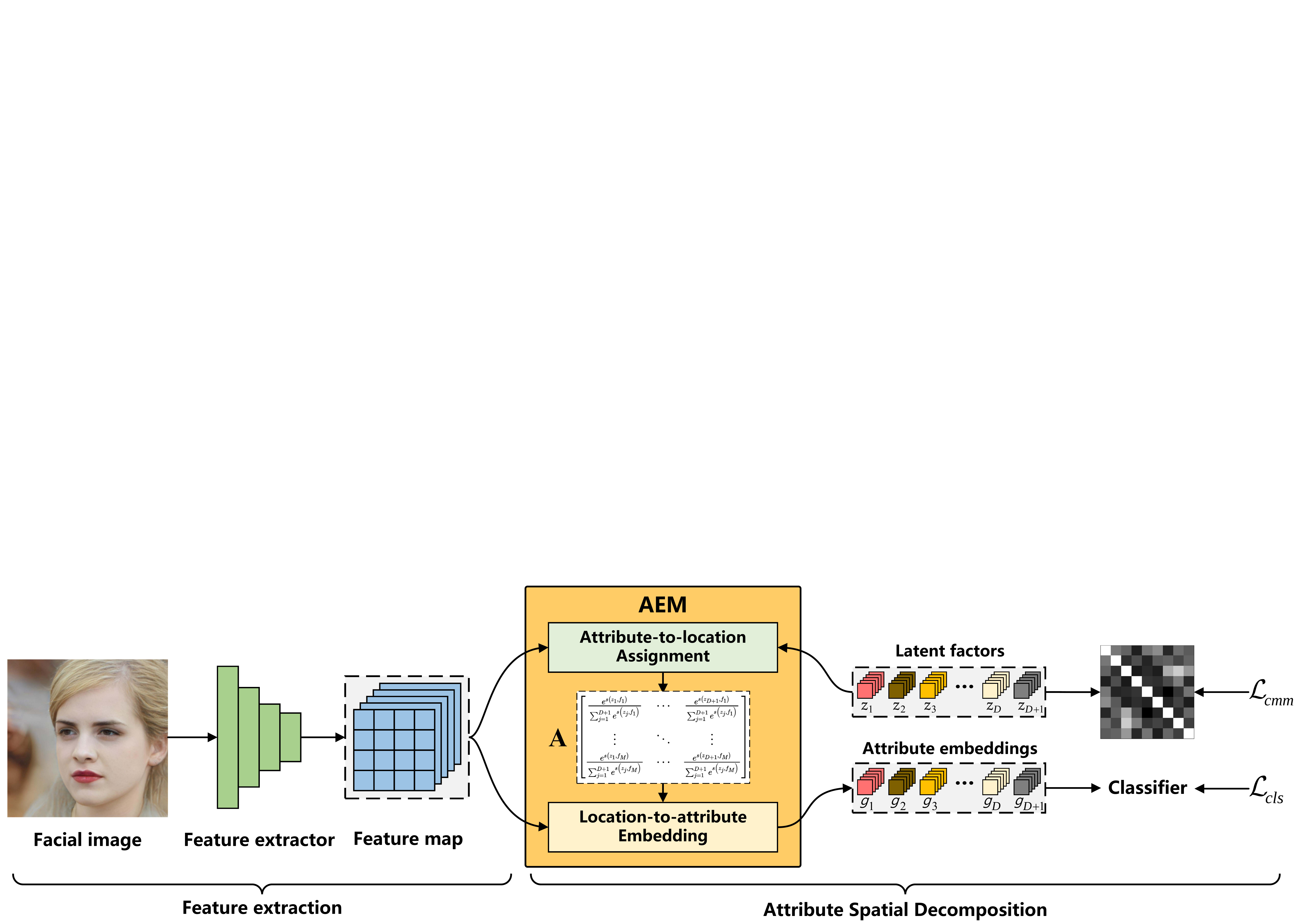}
	\caption{The overall framework of the proposed method consists of feature extraction and attribute spatial decomposition. 
		The feature map of the facial image is first extracted via the CNN-based feature extractor. Then, assignment-embedding module (AEM) is modeled to enable the procedure of ASD via attribute-to-location assignment and location-to-attribute embedding. Inspired by HFA \cite{Gong2013Hidden}, latent factors $z$ are introduced in attribute-to-location assignment to generate the assignment matrix $A$ between feature map and $z$. Location-to-attribute embedding integrates the attribute components from all locations guided with $A$ to represent the attribute embeddings $g$ of the entire image. Finally, cross entropy loss $\mathcal{L}_{cls}$ is conducted for the classifier, while we introduce correlation matrix minimization (CMM) $\mathcal{L}_{cmm}$ to constrain the correlation matrix $H$ of $z$. The correlations among $z$ can be reduced in order to strengthen the discriminability of $g$.}
	\label{fig:framework}
\end{figure*}

\section{Related Works}\label{sec:related}
\subsection{Facial attribute recognition}
With the significant success of deep learning in computer vision community, deep learning-based methods have emerged in the field of FAR. 
Liu {\itshape et al.} \cite{Liu2015Deep} introduce a large scale benchmark for FAR in the wild, termed as CelebFaces Attribute (CelebA). Meanwhile, a deep learning framework is proposed, which consists of two localization networks (LNet) and an attribute recognition network (ANet) for face localization and facial attribute prediction.
Rudd {\itshape et al.} \cite{Rudd2016MOON} cast facial attribute classification as a multi-task problem, in which the learning of each facial attribute is treated as a different task with joint optimization. 
Hand {\itshape et al.} \cite{hand2017attributes} design a multi-task deep convolutional network (MCNN) based on the relationships between facial attributes. Specifically, 40 facial attributes are first divided into 9 groups according to the facial spatial locations. Then, the shallow layers of MCNN are shared for all attribute groups, while the deep layers are independent belong to corresponding group. The attribute groups are treated as different tasks, and a multi-task learning paradigm is conducted to improve the final performances of FAR.
Han {\itshape et al.} \cite{Han2018Heterogeneous} estimate the facial attributes via representing the correlation and heterogeneity of attributes in a multi-task learning framework, in which the heterogeneity means the attribute data type, scale, and semantic meaning.
Cao {\itshape et al.} \cite{Cao2018Partially} introduce identification information to boost a partially shared multi-task convolutional neural network (PS-MCNN) for FAR. The attribute group-wise representations are shared hierarchically among five CNNs. 
He {\itshape et al.} \cite{He2018Harnessing} conduct a dual-path FAR network to leverage features from the original face images and facial abstraction images., where the facial abstraction images are generated by a generative adversarial network (GAN).
Chen {\itshape et al.} \cite{Chen2021Improving} introduce group attention learning and graph correlation learning to discover correlations among facial parts, while Multi-scale Group and Graph Network (MGG-Net) is proposed for FAR. 
Shu {\itshape et al.} \cite{Shu2021Learning} design three facial-related auxiliary tasks to learn spatial-semantic relationships between facial attributes, and then the spatial-semantic knowledge are transferred to FAR task.

The above mentioned methods often achieve the competitive performances relying on extra prior information.
However, the extra prior information, such as identifications, landmarks, and face parsing masks, might be not available,
resulting the limitation of these methods. In this paper, we focus on modeling FAR method without any extra prior information.

\subsection{Decomposition of facial components}
Face is composed of different characteristics physiologically, such as eyes, nose, and lips. 
Therefore, facial representation in facial analysis tasks can be intuitively decomposed into various task-related facial components. 
Here, we focus on the deep learning-based facial analysis methods with the decomposition.
Inspired by hidden factor analysis (HFA) \cite{Gong2013Hidden}, Wen {\itshape et al.} \cite{wen2016latent} propose a latent factor guided CNN (LF-CNN) to learn the age-invariant deep facial features for age-invariant face recognition. 
The observed facial features are formulated as a linear combination including the latent factors in terms of identification, age, noise and mean of facial feature. 
Li {\itshape et al.} \cite{li2019self} propose a twin-cycle autoencoder (TCAE) for action unit (AU) detection based on a self-supervised learning paradigm. 
The movements are factorized into AU-related and pose-related displacements among a pair of images, 
in which facial action-related movements can be disentangled from head motion-related movements.
Chen {\itshape et al.} \cite{chen2019semantic} propose a semantic component decomposition for face attribute manipulation.
Facial attribute is decomposed into multiple semantic components, each corresponds to a specific face region.
Alharbi {\itshape et al.} \cite{alharbi2020disentangled} exploit a grid structure-based noise injection to disentangle the latent space of facial image generation. 
The several aspects of disentanglement achieve fine-grained control over the generated images.
Cheng {\itshape et al.} \cite{cheng2021puregaze} propose a plug-and-play self-adversarial network to decompose facial feature. 
The network simultaneously removes entire image feature and preserves gaze-related feature,
improving the robust performance of facial gaze estimation.

The above methods of facial analysis decompose the face into global-level facial components, ignoring the ambiguity of facial components on the local locations. 
Here, we propose ASD which focuses on the decomposition of facial attribute components in the spatial dimension based on the spirit of HFA.
The attribute components are described via latent factors on the each spatial location to alleviate the spatial ambiguity. 
It is worth noting that the formulation of AEM is kind of similar to part localization module (PLM) in \cite{Zhao2019Recognizing}, but there is an obvious difference. 
PLM only conducts the learnable vectors to decompose attribute-related information from the feature map, 
ignoring to consider attribute-irrelated information.
In contrast, AEM represent the general formulation for the feature map, which is guided with latent factors including attributes, noise and mean of facial feature. Therefore, PLM can be regarded as a special case of AEM. 

\section{Attribute Spatial Decomposition}\label{sec:method}
\subsection{Overview}
ASD aims to represent facial attribute components for each spatial location of a facial image via latent factors, 
alleviating the spatial ambiguity without any extra prior information. 
AEM is a key idea in the procedure of ASD, 
which is adapted for deep learning-based FAR framework, as shown in Figure~\ref{fig:framework}.
Specifically, given a facial image $X$ whose feature map $f \in \mathbb{R}^{H \times W \times C}$ is first obtained via a CNN-based feature extractor $F$ as follows:
\begin{equation}
	f = F(X; \theta_{F}),
\end{equation}
where $\theta_{F}$ denotes the learnable parameters of feature extractor, $C$ is the number of channels and $H \times W$ is the resolution.
Then, AEM is modeled to decompose the various attribute components of $f$ in the spatial dimension via attribute-to-location assignment, 
while the attribute components are integrated via location-to-attribute embedding. 
The two operations of AEM can be formulated as follows:
\begin{equation}
	g = \psi(\phi(f;z), f),
\end{equation}
where $z$ represents the latent factors, $\phi$ and $\psi$ denote attribute-to-location assignment and location-to-attribute embedding, respectively. 
Finally, a classifier $P$ is conducted to project the integrated attribute embeddings $g$ to the label-space as follows:
\begin{equation}
	p= P(g; w_{P}, b_{P}),
\end{equation}
where $w_{P}$ and $b_{P}$ denote the learnable parameters of classifier.



\subsection{Assignment-Embedding Module}
AEM is in charge of decomposing the attribute components via attribute-to-location assignment and location-to-attribute embedding. 
Attribute-to-location assignment generates an assignment matrix via similarities between feature map and latent factors. The assignment matrix explicitly describes the magnitudes of attribute components on each spatial location,
which can be seen as a HFA-like formulation. It is worth noting that we generally formulate the feature map including the components of attributes, noise, and mean of facial feature.
Location-to-attribute embedding integrates attribute components from all locations guided with the assignment matrix to represent the attribute embeddings of the entire image. 

\subsubsection{Attribute-to-location Assignment}
\ 

The feature map $f$ is first flattened as $f'$ with the size of $M \times C$, where $M=H \times W$. 
Then, the HFA-like formulation of the $i$-th spatial location $\hat{f}_{i}$ can be defined as follows:
\begin{equation}\label{equ:hfa}
	\hat{f}_{i} =  \underbrace{\sum_{j=1}^{D} \alpha_{i,j}f_{i} + \alpha_{i,D+1}f_{i} + \bar{f},}_{\text{attributes, noise, and mean of feature}}
\end{equation}
where $\bar{f} \in \mathbb{R}^{C}$ is the mean of feature map, $f_{i} \in \mathbb{R}^{C}$ denotes the feature on the $i$-th spatial location of $f'$. 
The components of attributes, noise and mean of facial feature are combined linearly with assigning magnitudes.
Equation~\ref{equ:hfa} can explicitly describe the magnitude of each component in the spatial dimension, 
alleviating the spatial ambiguity of facial attributes formally.
$\alpha_{i,j}$ is the assigning magnitude in the $(i,j)$ position of assignment matrix $A$ which can be generated as follows:
\begin{equation}
	 \boldsymbol{\mathrm{A}} =\left[\begin{array}{ccc}
		\frac{e^{s(z_{1},f_{1})}}{\sum_{j=1}^{D+1} e^{s(z_{j},f_{1})}} & \cdots & \frac{e^{s(z_{D+1},f_{1})}}{\sum_{j=1}^{D+1} e^{s(z_{j},f_{1})}} \\
		\vdots & \ddots & \vdots \\
		\frac{e^{s(z_{1},f_{M})}}{\sum_{j=1}^{D+1} e^{s(z_{j},f_{M})}} & \cdots & \frac{e^{s(z_{D+1},f_{M})}}{\sum_{j=1}^{D+1} e^{s(z_{j},f_{M})}}
	\end{array}\right],
\end{equation}
where $\boldsymbol{\mathrm{A}} \in \mathbb{R}^{M \times (D+1)}$ denotes the assignment matrix. The assigning magnitude $\alpha$ is defined to describe the magnitude of components for each spatial location, including $D$ attributes and noise. 
$s$ denotes a cosine similarity between $f'$ and $z$, which is formulated as follows:
\begin{equation}\label{equ:sim}
	s(a,b) = \frac{a^{\mathsf{T}} b}{|a||b|},
\end{equation}	
where $| \cdot |$ is the modulo operation of vector.
$z_{j} \in \mathbb{R}^{C}$ is the $j$-th learnable latent factor of $z$, representing the $j$-th attribute in the latent space.
$z_{D+1}$ denotes the latent noise factor which is an additional term in the formulation. 
We argue that $z_{D+1}$ can guide an explicit assignment of the attribute-irrelated components, 
enhancing the discriminability of the attribute-related.

\subsubsection{Location-to-attribute Embedding}
\ 

After attribute-to-location assignment, attribute components can be extracted from each spatial location explicitly via assigning magnitudes. 
Then, the attribute components from all locations are integrated to attribute embeddings via location-to-attribute embedding as follows:
\begin{equation}
	g_{j} = \sum_{i=1}^{M} \alpha_{i,j}f_{i} + \bar{f},
\end{equation}
where $g_{j} \in \mathbb{R}^{C}$ is the $j$-th attribute embedding. 
Furthermore, the entire attribute embeddings $g$ can be simplified with matrix operations as follows:
\begin{equation}
	g = \boldsymbol{\mathrm{A}}^{\mathsf{T}} \! f' +  [ \, \bar{f} \, ]_{(D+1) \times C,}
\end{equation}
where $[ \, \bar{f} \, ]_{(D+1) \times C} \in \mathbb{R}^{(D+1) \times C}$ denotes a matrix whose entities are $\bar{f}$.
To formulate the attribute embeddings generally, we preserve the noise embedding as the $(D+1)$-th embedding. Finally, the two operations of AEM are summarized as the pseudo code in \mbox{Algorithm \ref{alg:aem}}.

\begin{algorithm}[t]
	\centering
	\caption{\small{Pseudo code of assignment-embedding module (AEM) in a PyTorch-like style.}}
	\label{alg:aem}
	\definecolor{codeblue}{rgb}{0.25,0.5,0.5}
	
	\lstset{
		backgroundcolor=\color{white},
		basicstyle=\fontsize{7.2pt}{7.2pt}\ttfamily\selectfont,
		columns=fullflexible,
		breaklines=true,
		captionpos=b,
		commentstyle=\fontsize{7.2pt}{7.2pt}\color{codeblue},
		keywordstyle=\fontsize{7.2pt}{7.2pt}\color{black},
	}
	\begin{lstlisting}[language=python]
# In: Feature map f. 
# Out: Attribute embeddings g.

class AEM(nn.Module):
def __init__(self, D, C):
   self.z = nn.Parameter(torch.rand((D+1, C)))

def forward(self, f):
   # To simplification, the dimension of batchsize is omitted.
   W, H, C = f.size()
   # f is flattened as [M, C].
   f = f.view(W*H, C)
   # Attribute-to-location assignment.
   norm_z, norm_f = F.normalize(self.z), F.normalize(f)
   A = torch.exp(norm_f @ norm_z.t())
   A /= torch.sum(A)
   # Location-to-attribute embedding.
   g = A.t() @ f + torch.mean(f)
   return g
	\end{lstlisting}
\end{algorithm}

\subsection{Correlation Matrix Minimization}

Since the high correlation among latent factors would weaken the distinctions of assignment magnitudes in $A$,
we introduce a regularization, termed as correlation matrix minimization (CMM), to reduce the correlation among $z$. The correlation matrix of $z$ can be obtained as follows:
\begin{equation}
	\boldsymbol{\mathrm{H}} = \left[\begin{array}{ccc}
		s(z_{1},z_{1}) & \cdots & s(z_{1},z_{D+1}) \\
		\vdots & \ddots & \vdots \\
		s(z_{D+1},z_{1}) & \cdots & s(z_{D+1},z_{D+1})
	\end{array}\right],
\end{equation}
where $s$ is presented in Equation~\ref{equ:sim}. 
Then, CMM can be formulated as follows:
\begin{equation}
	\mathcal{L}_{cmm} = \sum_{i=1}^{D+1} (1 - s(z_{i},z_{i}))^{2} + \sum_{i=1}^{D+1} \sum_{ j=1, i \neq j }^{D+1}   \!\! s^{2}(z_{i},z_{j}),
\end{equation}
where the first term can be omitted obviously, since 
\begin{math}
	s(z_{i},z_{i}) \triangleq 1
\end{math}, and
the second term enforces the off-diagonal elements of correlation matrix to 0, 
decorrelating the latent factors explicitly.

\subsection{Classification and Loss Function}
For each attribute embedding, we cast the attribute predictions as multiple binary classification tasks.
The $j$-th representation of attribute embedding is projected to a logit value linearly, 
and then a sigmoid function $\sigma$ is utilized to convert the logit value as a probability $p_{j}$,
which can be formulated as follows:
\begin{equation}
	p_{j} = \sigma (w^{\mathsf{T}}_{P_{j}} g_{j} + b_{P_{j}}),
\end{equation}
where $w_{P_{j}}$ and $b_{P_{j}}$ refer to the weight vector and bias, respectively. 
The final prediction $p$ is generated via concatenating the probabilities of attribute embeddings as follows:
\begin{equation}
	p = \big[p_{1}, p_{2}, \dots , p_{D}, p_{D+1}\big],
\end{equation}
where $p \in \mathbb{H}^{D+1}$ is the prediction of $D$ attributes and noise, where $\mathbb{H}$ represents Hamming space.

The loss function of the classification is cross entropy loss which can be formulated as follows:
\begin{equation}
	\mathcal{L}_{cls} = -  \frac{1}{D+1} \sum^{D+1}_{j=1} y_{j} \log p_{j},
\end{equation}
where $y_{j}$ is the label of the $j$-th attributes. $y_{D+1}$ is set to 0, since the label of noise is not available. 
The final loss function can be defined as follows:
\begin{equation}
	\mathcal{L} = \mathcal{L}_{cls} + \gamma \mathcal{L}_{cmm},
\end{equation}
where $\gamma$ is a hyper-parameter to balance the importance between the classification and the constraint of latent factors.



\section{Experiment}\label{sec:experiment}
\subsection{Datasets}
We conduct experiments on two public facial attribute datasets including CelebA \cite{Liu2015Deep} and LFWA \cite{Liu2015Deep}, which is widely used to evaluate the method of FAR. 

CelebA is a large-scale facial attribute dataset containing 202,599 facial images divided into 3 subsets in terms of training, validation, and testing.
The numbers of facial images for 3 subsets are 162,770, 19,867, and 19,962, respectively.
The facial images are annotated with 40 attribute labels.

LFWA is another popular facial attribute dataset composed of 13,143 facial images with
6,263 for training, 2,800 for validation, and 4,080 for testing.
The facial images are also annotated with the same attribute labels as CelebA.

In the experiments, we follow the protocol of CelebA and LFWA that the default training set is used to train our method, 
while the performance of our method is evaluated on the default testing set.

\subsection{Implementation}
The experiments are conducted on a work station with NVIDIA RTX 2080Ti GPUs. 
The proposed method is implemented based on PyTorch deep learning framework.
The backbone of feature extractor is ResNet50 \cite{He2016Deep} pre-trained on ImageNet \cite{Deng2009ImageNet},
whose capability of feature extraction has been verified in many computer vision tasks \cite{Hu2020Efficient, Basu2022Do, Vente2022Automated, Zhang2022Character}.
Following \cite{Gong2013Hidden, wen2016latent}, 
the latent factors $z$ are initialized with the distribution of $\mathcal{N}(0, I)$.
In the training phase, Adam \cite{kingma2015adam} is used to optimize the learnable parameters with a weight decay of 5e-4. 
The total epochs are 80 and the initial learning rate is 3e-4, 
while the learning rate is reduced with the decay ratio of 0.1 after every 20 epochs. 
The value of $\gamma$ is set to 2e-2 empirically.
The input images are scaled as $224 \times 224$ and random flip is used as data argumentation.
Particularly, the sizes of batches for training on CelebA and LFWA are 64 and 32, respectively.

\begin{figure*}[!t]
	\centering
	\includegraphics[width=\linewidth]{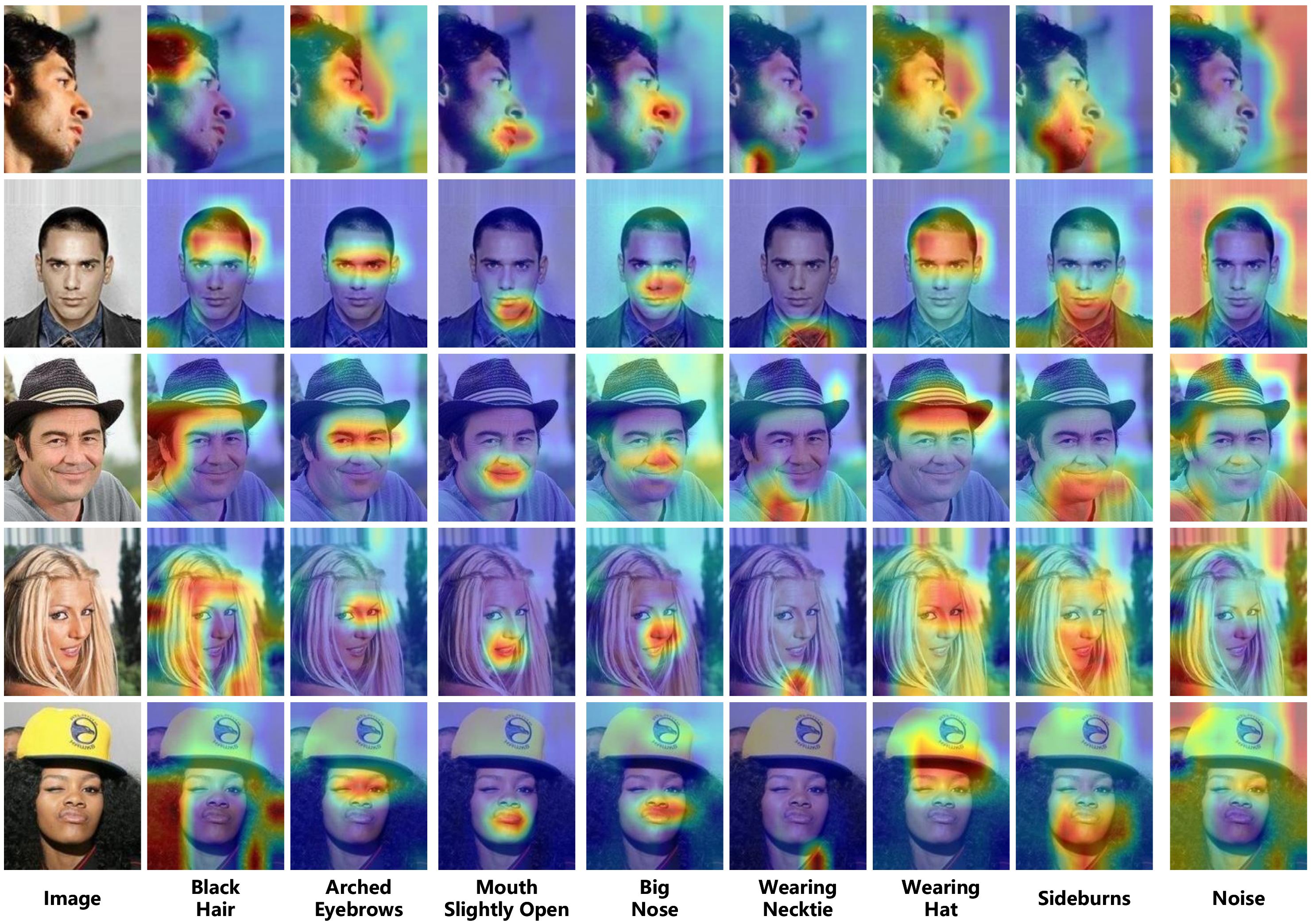} 
	\caption{The visualization of assignment matrix $A$ for some traditional attributes, such as {\itshape Black Hair}, {\itshape Arched Eyebrows}, {\itshape Mouth Slightly Open}, {\itshape Big Nose}, {\itshape Wearing Necktie}, {\itshape Wearing Hat}, and {\itshape Sideburns}. Specifically, $A$ is rearranged with the size of $H \times W \times (D+1)$, and the assigning magnitudes are visualized via heatmaps to represent the corresponding attribute components in the spatial dimension. It can be observed that the spatial locations belong to the corresponding facial attributes are highlighted obviously. The capability of ASD to decompose the attribute components is verified visually. Furthermore, we illustrate the assigned noise components in the last column. The attribute-irrelated regions are highlighted via heatmaps, which is argued the insight of introducing the noise embedding $z_{D+1}$. 
	}
	\label{fig:heatmap}
\end{figure*}

\subsection{Ablation studies}
To evaluate the effectiveness of the proposed method comprehensively, 
we design the ablation studies in terms of ASD, AEM, and CMM.

\subsubsection{The effectiveness of ASD for different feature extractors}
\
\begin{table}
	\centering
	\caption{Ablation analysis of ASD for the different feature extractors. 
		``w/ ASD'' represents the method constructed with attribute spatial decomposition.}
	\label{tab:abla_backbone}
	\begin{tabular}{cccc}
		\toprule
		Backbone  & w/ ASD & CelebA (\%)& LFWA (\%)\\
		\midrule
		ResNet18  &        & 90.93      & 85.75    \\
		ResNet18  &    \checkmark    & 91.67      & 86.68    \\
		ResNet50  &        & 91.39      & 86.47    \\
		ResNet50  &    \checkmark   & 92.22      & 87.43    \\
		ResNeXt50 &        & 91.63      & 86.78    \\
		ResNeXt50 &   \checkmark   & 92.21      & 87.44   \\
		\bottomrule
	\end{tabular}
\end{table}
To clarify the effectiveness of ASD, we conduct the ablation experiments based on the different feature extractors.
Three popular CNNs for FAR are introduced, 
including ResNet18 \cite{He2016Deep}, ResNet50 \cite{He2016Deep}, and ResNeXt50 \cite{xie2017aggregated}. 
Specifically, two versions of FAR method for corresponding feature extractors are constructed. 
The first version of the method is implemented without ASD,  
in which the global pooled feature is directly used as the input of the fully connected layer-based classifier.
In contrast, the second version of method is designed with ASD.
The results of the different feature extractors-based methods in terms of average accuracy are reported in Table~\ref{tab:abla_backbone}.
We observe that ASD improves the performances of FAR obviously.
For CelebA, the average accuracies of the methods based on various backbones are improved about 0.7\%, 0.9\%, and 0.6\%, respectively.
For LFWA, the improvements caused via ASD are 0.9\%, 0.9\%, and 0.7\%, respectively.
The experimental results demonstrate the effectiveness of ASD whose superiority is unaffected by the different feature extractors.
Moreover, to interpret ASD intuitively, we visualize the assignment matrix $A$ which is the key idea of ASD, as shown in Figure~\ref{fig:heatmap}. 
The heatmaps highlight the corresponding attribute-related regions evidently.
It argues that the attribute components can be described via ASD in the spatial dimension.
The spatial ambiguity of facial attributes is alleviated, thereby improving the performance of FAR effectively.
To consider the trade-off between efficiency and performance, we implement ASD based on ResNet50 in the subsequent experiments.

\begin{table}
	\centering
	\caption{Ablation analysis of $z_{D+1}$ and $\bar{f}$. 
		``w/o $z_{D+1}$'' and ``w/o $\bar{f}$'' denote AEM without latent noise factor and mean of feature, respectively.}
	\label{tab:abla_latent}
	\begin{tabular}{cccc}
		\toprule
		w/o $z_{D+1}$ & w/o $\bar{f}$ & CelebA (\%) & LFWA (\%)\\
		\midrule
		\checkmark & \checkmark  & 91.76      & 86.98    \\ 
		&  \checkmark & 91.91      & 87.23    \\
		\checkmark &   & 91.93     & 87.22    \\
		\bottomrule
	\end{tabular}
\end{table}

\begin{table}
	\centering
	\caption{Ablation analysis of CMM. 
		``w/ CMM'' denotes AEM correlation matrix minimization.}
	\label{tab:abla_cmm}
	\begin{tabular}{ccc}
		\toprule
		w/ CMM & CelebA (\%) & LFWA (\%)\\
		\midrule
		& 92.03      & 87.27    \\ 
		\checkmark    & 92.22     & 87.43    \\
		\bottomrule
	\end{tabular}
\end{table}

\subsubsection{Latent noise factor and mean of feature in AEM}
\
%

To investigate the impact of latent noise factor $z_{D+1}$ and mean of feature $\bar{f}$,
we design the methods based on ResNet50 without $z_{D+1}$ and $\bar{f}$, separately.
Meanwhile, the method without both $z_{D+1}$ and $\bar{f}$ is constructed, termed as vanilla AEM.
As shown in Table~\ref{tab:abla_latent}, 
the performance of the method including $z_{D+1}$ is higher than the vanilla AEM,
while the formulation of $\bar{f}$ in the AEM also improves the overall performance.

\subsubsection{The effectiveness of CMM}
\

To validate the effectiveness of CMM, we further train the method based on ResNet50 without CMM.
The quantitative results are reported in Table~\ref{tab:abla_cmm}. 
The method with CMM achieves a higher accuracy that about 0.2\% increases on both CelebA and LFWA.
It argues that the decorrelation among latent factors can enhance the discriminability of $g$ to improve the overall performance of ASD.

\subsection{Comparison with State-of-the-art Methods}

\begin{table}
	\centering
	\caption{Comparison of our method with the state-of-the-art methods on CelebA and LFWA. 
		``w/ EP'' and ``w/ IP'' denote the method with explicit prior information and implicit prior information, respectively. The best and the second best performances of average accuracy (\%) are marked with {\itshape \color{red}red} and {\itshape \color{blue}blue}, respectively. ``-'' represents the result is not provided by corresponding paper.
	}
	\label{tab:sota}
	\begin{tabular}{ccccc}
		\toprule
		Method   & w/ EP & w/ IP & CelebA (\%) & LFWA (\%)  \\
		\midrule
		AFFAIR \cite{li2018landmark}   &  \checkmark   &        & 91.45  & 86.13 \\
		PS-MCNN \cite{Cao2018Partially}  &  \checkmark   &  \checkmark & {\color{red}92.98}  & {\color{red}87.36} \\
		HSA \cite{He2018Harnessing}      &   \checkmark   &             & 91.81  & 85.20  \\
		DMM \cite{mao2020deep}     &    \checkmark     &  \checkmark  & 91.70   & 86.56 \\
		SlimCNN \cite{sharma2020slim} &               &   \checkmark & 91.24  & 76.02 \\
		SSPL \cite{Shu2021Learning}    &   \checkmark      &           & 91.77  & 86.53 \\
		MGG-Net \cite{Chen2021Improving} &           &   \checkmark    & 92.00     & 87.20  \\
		HFE \cite{Yang2020Hierarchical} & & \checkmark & 91.24 & - \\
		ASD (Ours)    &                        &                        &    {\color{blue}92.22}    &   {\color{blue}87.43}    \\
		\midrule
		CSN \cite{Zhao2019Recognizing}     &                        &                        & 91.80   & -   \\
		TResNetM \cite{ridnik2021tresnet}&                        &                        &   {\color{blue}91.72}     &   {\color{blue}86.67}    \\
		ASD (Ours)    &                        &                        &    {\color{red}92.22}    &    {\color{red}87.43}   \\
		\bottomrule
	\end{tabular}
\end{table}

%
We compare our method with 10 state-of-the-art methods lately reported on CelebA and LFWA, 
which can be categorized into as prior-based and prior-free methods.
The prior-based methods include ARRAIR \cite{li2018landmark}, PS-MCNN \cite{Cao2018Partially}, HSA \cite{He2018Harnessing}, DMM \cite{mao2020deep}, SlimCNN \cite{sharma2020slim}, SSPL  \cite{Shu2021Learning},  MGG-Net \cite{Chen2021Improving}, and HFE\cite{Yang2020Hierarchical}, where prior information, such as landmarks, face parsing masks, and attribute prior embeddings, are utilized in the training phase.
CSN \cite{Zhao2019Recognizing} and TResNetM \cite{ridnik2021tresnet} are general prior-free methods for multi-label classification task. 
Here, we implement TResNetM on CelebA and LFWA with the training scheme provided in \cite{ridnik2021tresnet}, such as optimizer, learning rate, and training epochs. 
Besides \cite{ridnik2021tresnet}, we list the experimental results of other methods reported from corresponding papers, as shown in Table~\ref{tab:sota}.

For the comparison with prior-based methods, 
ASD achieves the competitive performances than state-of-the-art methods on both CelebA and LFWA.
It can be observed that there is still a performance gap between ASD and PS-MCNN.
However, we argue the superiority of ASD is obvious. 
First, the architecture of ASD is simpler. PS-MCNN consists of five CNN-based networks whose initializations are pre-trained on face recognition tasks, separately. 
Such stage-based training procedure inevitably increases the complicacy of the implementation.
In contrast, ASD is composed of a single CNN-based network and AEM, which can be trained via an end-to-end paradigm. 
The another advantage of ASD is that the performance of ASD is not relied on extra prior information. 
In \cite{Cao2018Partially}, the performance of PS-MCNN without identified information and prior attribute groups would be deteriorated, 
where the average accuracy of PS-MCNN on CelebA is reduced from 92.98\% to 91.15\%.

For the comparison with prior-free methods, the performance of ASD is superior than CSN and TResNetM.
The reason for the low performance of TResNetM might be that the larger batches should be conducted, like the setting in \cite{ridnik2021tresnet}. 
To come up with a fair comparison, we gradually adopt the various sizes of batches for TResNetM trained on CelebA. 
The experimental results are listed in Table~\ref{tab:sota_batch}. 
It can be seen that the improvements of TResNetM caused by increasing the size of batches are slight,
while there is a significant growth of GPU memory costs in the training phase. 
Here, ASD achieves a good trade-off between average accuracy and computational burdens, 
implying the architecture of ASD without bells and whistles.

\begin{table}
	\centering
	\caption{Comparison of our method with TResNetM for various sizes of batches on CelebA. }
	\label{tab:sota_batch}
	\begin{tabular}{cccc}
		\toprule
		Method & batches & GPU memory & CelebA (\%) \\
		\midrule
		TResNetM & 64 & 4,200 MiB  & 91.72       \\
		TResNetM & 128 &  6,300 MiB & 91.97         \\
		TResNetM & 256 & 12,100 MiB  & 92.23        \\
		ASD (Ours) & 64 & 5,600 MiB  & 92.22       \\
		\bottomrule
	\end{tabular}
\end{table}

\begin{table}
	\caption{Performance with limited training data on CelebA.
		0.2\%, 0.5\%, 1\%, and 2\% are the different ratios of the limited training sets including 325, 832, 1,627, and 3,255 training samples, respectively.
		The best performance of average accuracy (\%) is marked with {\itshape \color{red}red}. }
	\label{tab:limited}
	\begin{tabular}{ccccc}
		\toprule
		Method & \multicolumn{4}{c}{CelebA (\%)}      \\
		ratios  & 0.2\% & 0.5\% & 1\%   & 2\%   \\
		numbers of training samples & 325 & 832 & 1627  & 3255 \\
		\midrule
		DeepCluster \cite{caron2018deep}             & 83.21  & 86.13  & 87.46 & 88.86 \\
		JigsawPuzzle \cite{noroozi2016unsupervised}           & 82.88  & 84.71  & 86.25 & 87.77 \\
		Rot \cite{gidaris2018unsupervised}                    & 83.25  & 86.51  & 87.67 & 88.82 \\
		FixMatch \cite{sohn2020fixmatch}              & 80.22  & 84.19  & 85.77 & 86.14 \\
		VAT \cite{miyato2018virtual}                    & 81.44  & 84.02  & 86.30  & 87.28 \\
		SSPL \cite{Shu2021Learning}                   & 86.67  & 88.05  & 88.84 & 89.58 \\
		ASD (Ours)              &  {\color{red}87.82} & {\color{red}89.19}  & {\color{red}89.90}  & {\color{red}90.23} \\
		\bottomrule
	\end{tabular}
\end{table}

\subsection{Performance with Limited Training Data}

The limited training data case is a latest challenge for FAR,
where only a small ratios of training samples can be utilized in the training phase.
To investigate the performance of ASD with limited training data, 
ASD is compared with other methods including DeepCluster \cite{caron2018deep},
JigsawPuzzle \cite{noroozi2016unsupervised}, Rotation \cite{gidaris2018unsupervised}, FixMatch \cite{sohn2020fixmatch}, 
VAT \cite{miyato2018virtual}, and SSPL \cite{Shu2021Learning}.

Following \cite{Shu2021Learning}, 
we select the samples randomly from the training set with various small ratios (0.1\%, 0.2\%, 1\%, and 2\%), 
constructing the limited training set for ASD.
The experimental results of each ratio are the mean of 10 times, 
while the limited training sets are repartitioned in each trial.
The performances of ASD with the different ratios of limited training sets are reported in Table~\ref{tab:limited}.
The experimental results show the the superior performance of ASD comparing with other methods,
revealing the reliable capability of ASD with the limited training data case for FAR. 


\section{Conclusion}\label{sec:conclusion}

In this paper, we propose a novel prior-free method for facial attribute recognition (FAR), 
termed as attribute spatial decomposition (ASD), 
mitigating the spatial ambiguity of facial attributes formally without any extra prior information.
Assignment-embedding module (AEM) is modeled to enable ASD via latent factors, 
while correlation matrix minimization (CMM) is introduced to improve the discriminability of decomposed attribute embeddings.
Experimental results demonstrate that 
ASD achieves the competitive performances without any extra prior information compared with state-of-the-art prior-based methods. Furthermore, the reliable performance of ASD for the case of limited training data is verified.

\bibliographystyle{IEEEtran} 
\bibliography{sample_sigconf}

%
\begin{IEEEbiography}[{\includegraphics[width=1in,height=1.25in,clip]{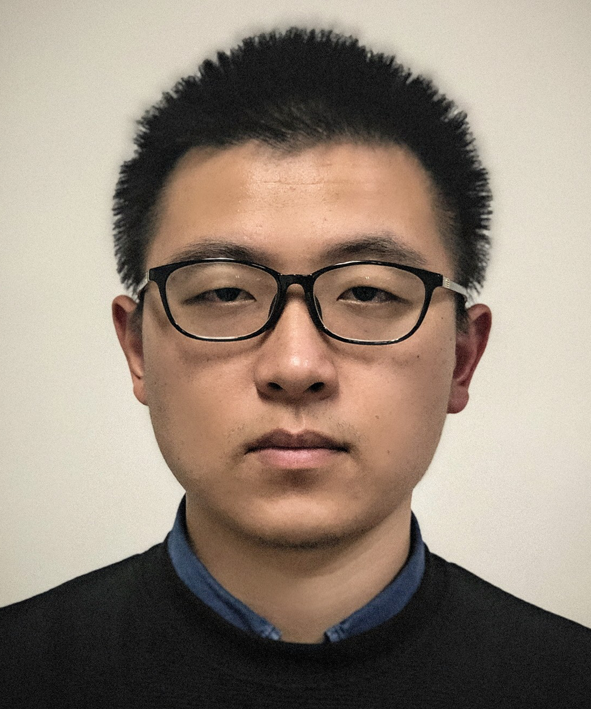}}] {Chuanfei Hu} received the M.S. degree from the School of Optical-Electrical and Computer Engineering, University of Shanghai for Science and Technology, Shanghai, China. He is currently pursuing the Ph.D. degree with the School of Automation, Southeast University, Nanjing, China, in 2021. His research interests include deep learning, attribute learning, and affective computing.
\end{IEEEbiography}

\begin{IEEEbiography}[{\includegraphics[width=1in,height=1.25in,clip]{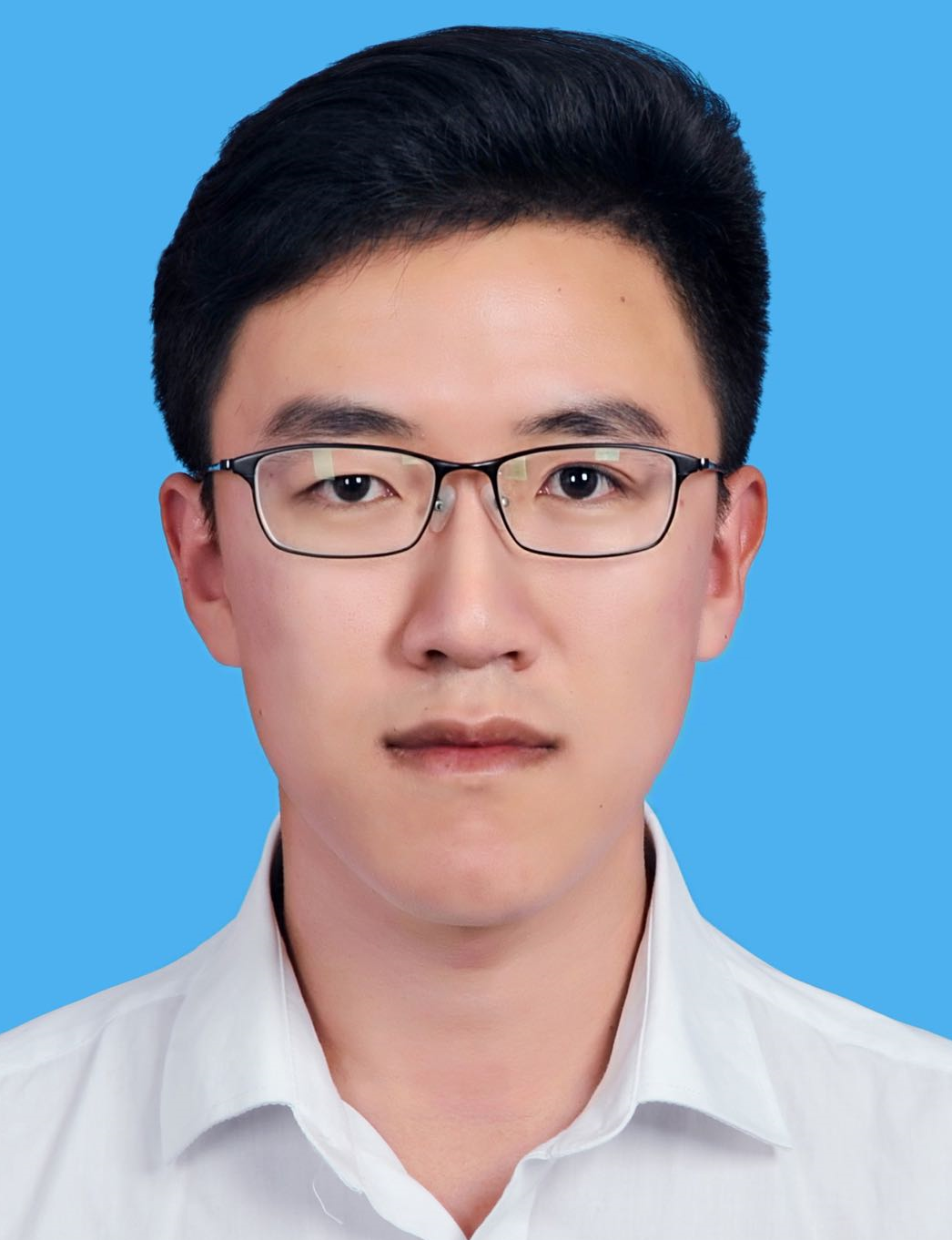}}] {Hang Shao} received the M.S. degree from the School of Optical-Electrical and Computer Engineering, University of Shanghai for Science and Technology, Shanghai, China. He is currently pursuing the Ph.D. degree with School of Computer Science and Engineering, Nanjing University of Science and Technology, Nanjing, China, in 2021. His research interests include deep learning and representation learning. 
\end{IEEEbiography}

\begin{IEEEbiography}[{\includegraphics[width=1in,height=1.25in,clip]{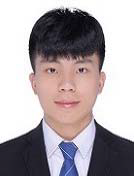}}] {Bo Dong} received the B.S. degree from the School of Optical-Electrical and Computer Engineering, University of Shanghai for Science and Technology, Shanghai, China. He is currently pursuing the M.S. degree in biomedical engineering with Zhejiang University, Zhejiang, China, in 2021. His main research interests include computer version and medical image analysis.
\end{IEEEbiography}

\begin{IEEEbiography}[{\includegraphics[width=1in,height=1.25in,clip]{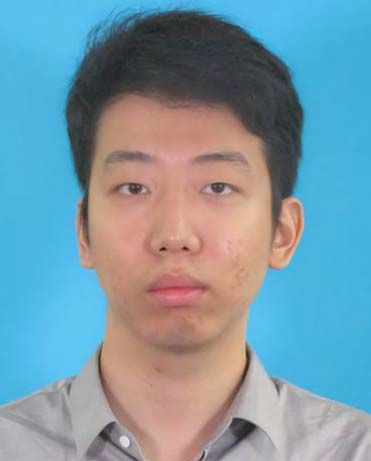}}] {Zhe Wang} received the M.S. degree from the School of Electrical and Electronic Engineering, Tianjin University of Technology, Tianjin, China. He is currently pursuing the Ph.D. degree with the School of Optical-Electrical and Computer Engineering, University of Shanghai for Science and Technology, Shanghai, China, in 2020. His research interests include affective computing, signal processing, and brain-computer interface.
\end{IEEEbiography}

\begin{IEEEbiography}[{\includegraphics[width=1in,height=1.25in,clip]{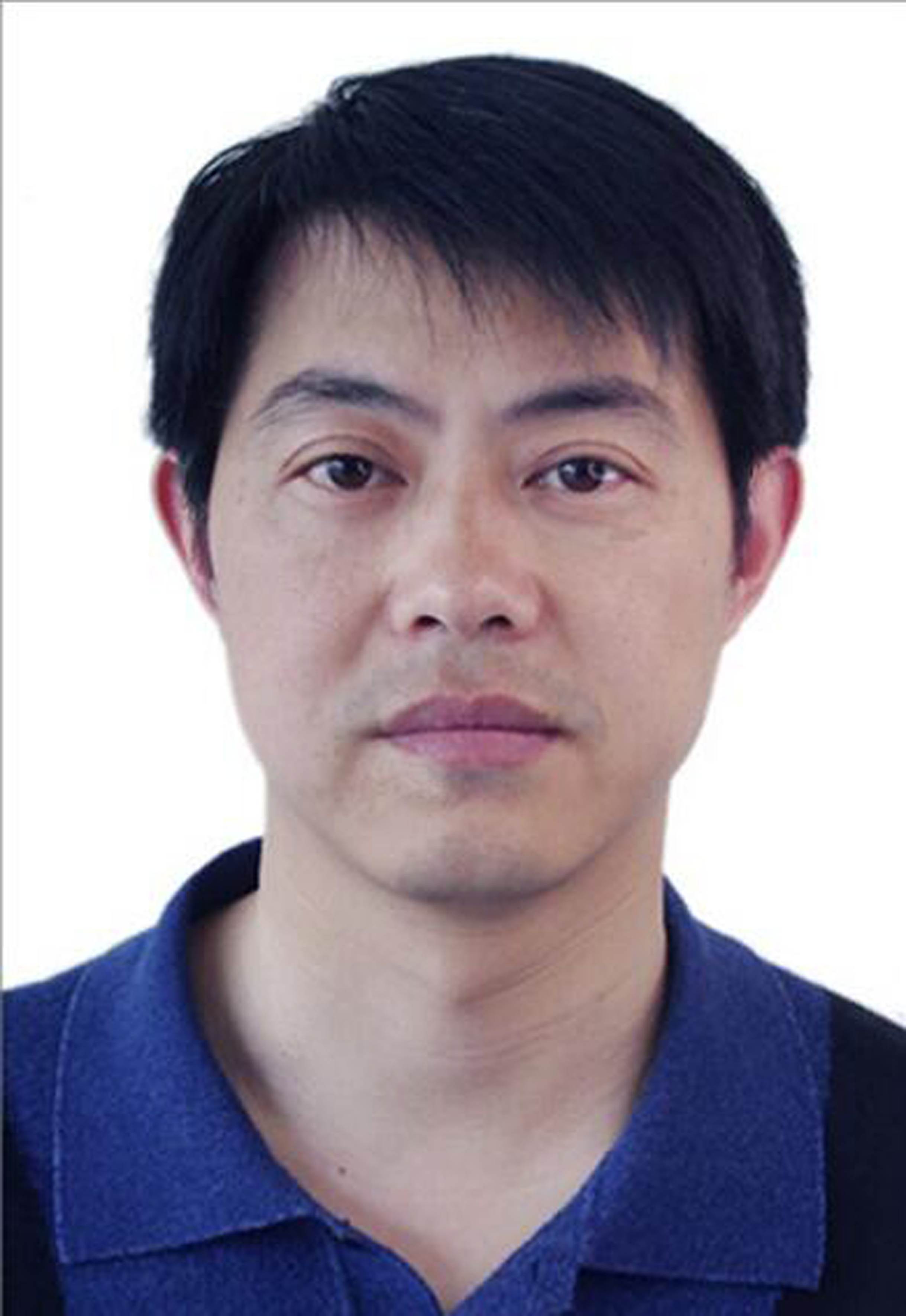}}] {Yongxiong Wang} received the B.S. degree in engineering mechanics from Harbin Engineering University, Harbin, China, and the M.S. and Ph.D. degrees in control science and engineering from Shanghai Jiao Tong University, Shanghai, China, in 1991. He is currently a Professor with the School of Optical-Electrical and Computer Engineering, University of Shanghai for Science and Technology, Shanghai. His research interests include computer vision, affective computing, and intelligent robot.
\end{IEEEbiography}

\end{document}